\documentclass{article}

\usepackage{microtype}
\usepackage{graphicx}
\usepackage{subcaption}
\usepackage{booktabs} 

\usepackage{amsfonts}       
\usepackage{nicefrac}       
\usepackage{amsmath}
\usepackage{comment} 
\usepackage{pdflscape}
\usepackage{soul}

\usepackage{multirow}

\usepackage{hyperref}

\usepackage[<package options>]{movie15}

\usepackage[accepted]{icml2019}

\icmltitlerunning{Relevance Factor VAE}

\begin{document}

\twocolumn[
\icmltitle{Relevance Factor VAE: 
Learning and Identifying Disentangled Factors}



\icmlsetsymbol{equal}{*}

\begin{icmlauthorlist}
\icmlauthor{Minyoung Kim}{ru,snut}
\icmlauthor{Yuting Wang}{ru}
\icmlauthor{Pritish Sahu}{ru}
\icmlauthor{Vladimir Pavlovic}{ru}
\end{icmlauthorlist}

\icmlaffiliation{ru}{Dept. of Computer Science, Rutgers University, NJ, USA}
\icmlaffiliation{snut}{Dept. of Electronic Engineering, Seoul National University of Science $\&$ Technology, Seoul, South Korea}

\icmlcorrespondingauthor{Vladimir Pavlovic}{vladimir@cs.rutgers.edu}

\icmlkeywords{Machine Learning, ICML}

\vskip 0.3in
]



\printAffiliationsAndNotice{}  

\begin{abstract}
We propose a novel VAE-based deep auto-encoder model that can learn disentangled latent representations in a fully unsupervised manner, endowed with the ability to identify all meaningful sources of variation and their cardinality. Our model, dubbed {\em Relevance-Factor-VAE}, leverages the total correlation (TC) in the latent space to achieve the disentanglement goal, 
but also addresses the key issue of existing approaches which cannot distinguish between meaningful and nuisance factors of latent variation, often the source of considerable degradation in disentanglement performance. 
We tackle this issue by introducing the so-called relevance indicator variables that can be automatically learned from data, together with the VAE parameters. Our model effectively focuses the TC loss onto the relevant factors only by tolerating large prior KL divergences, a desideratum  justified by our semi-parametric theoretical analysis. Using a suite of disentanglement metrics, including a newly proposed one, as well as qualitative evidence, we demonstrate that our model outperforms existing methods across several challenging benchmark datasets. 
\end{abstract}

\section{Introduction}\label{sec:intro}

Identifying the true underlying factors or sources that explain the variability and structure of data is a key problem in machine learning.  Understanding ``how'' decisions are made is intimately tied to finding the representation that most succinctly explains the data. Such representations make it easier to extract useful information for constructing and interpreting predictive models,  important when the prediction goals only partially reflect the underlying structure of the data.  Learning this representation of data is also fundamental to understanding the complex observations in many applications~\cite{bengio_disent,lake16}. Latent variable models are the main tools for this task, thanks to their ability to principally model compact representation of observed data.  Deep learning architectures, embodied in latent variable models such as the deep variational auto-encoder (VAE)~\cite{vae14}, have extended this ability to modeling of complex nonlinear dependencies among latent factors and the ambient data.  

One of the highly desired requirements in learning the latent representation is the {\em disentanglement} of  factors: each latent variable shall be exclusively responsible for the variation of a unique aspect in the observed data~\cite{bengio_disent}. For example, for facial images, changing the value of one latent variable while fixing the others should result in variation of the azimuth pose alone while the other aspects, such as the subject ID, gender, skin color, or the facial expression, remain intact. Conventional data likelihood maximization learning, inherently adopted in the VAE, does not explicitly aim to achieve this goal.  This has given rise to several recent efforts aimed at amending the learning algorithms with the goal of constructing highly disentangled representations. While some of the approaches are (semi-)supervised, exploiting known factors of variations such as data labels~\cite{sup_reed, sup_yang, sup_kulkarni, sup_whitney}, in this paper we tackle the unsupervised setup: learning disentangled representations from unlabeled data alone.

Previous works on unsupervised disentangled representation learning have empirically demonstrated that one can achieve improved disentanglement by over-emphasizing: either the latent prior-posterior matching~\cite{aae16,beta_vae17}, or the mutual information between observed and latent variables~\cite{infogan16}. More recently, the idea of minimizing the total correlation (TC) was introduced, leading to significant improvements over the non-TC methods. Factor-VAE~\cite{factor_vae18} has introduced a percipient way of optimizing the TC, 
followed by alternative strategies such as ~\cite{tcvae}. 

Despite their potential, such models have not been able to address the key issue of distinguishing meaningful latent factors from the nuisance ones, which can lead to potentially significant degradation of disentanglement performance.  This is a consequence of the quintessential learning setting adopted here: to learn the disentangled representation, one will commonly choose a latent dimension sufficiently large to encompass both the major as well as the minor, nuisance, modes of variation. 
However, most existing methods disregard this fact, applying a homogeneous learning loss (e.g., the TC loss) to major and nuisance factors alike. 

The main goal of this paper is to address this issue by introducing the so-called relevance indicator variables that can be automatically learned from data together with the rest of VAE parameters.  Our model effectively takes into account the TC loss by focusing it only on relevant, disentangled factors, tolerating large prior divergence of these factors from those a priori specified in the nuisance model, while simultaneously attempting to identify the noise factors with small divergence from the same nuisance priors, 
an approach introduced in~\autoref{sec:rfvae} and fully justified by detailed theoretical analysis in the Supplement.
This enables automatic identification and learning of true factors, even in instances when the total number of such factors is not explicitly known.  We empirically demonstrate, through extensive empirical evaluations on several benchmark datasets, that our model significantly outperforms existing methods both quantitatively, across diverse disentanglement metrics, as well as qualitatively.  To that end, we also introduce a new disentanglement evaluation score, which shows improved agreement with qualitative assessment of disentangling models.

\section{Background}\label{sec:background}

\subsection{Notation and VAE~\cite{vae14}}

Let ${\bf x}$ be our observation (e.g., an image) and ${\bf z}\in \mathbb{R}^d$ be the underlying latent vector of ${\bf x}$. 
To represent the observed data, the variational auto-encoder (VAE) specifically defines a probabilistic model that incorporates the deep functional modeling, namely
\begin{eqnarray}
\vspace{-0.5em}
p({\bf z}) &=& \mathcal{N}({\bf z}; {\bf 0}, {\bf I}), \label{eq:p_z} \\
p({\bf x}|{\bf z}) &=& p({\bf x}; \theta({\bf z})), \label{eq:p_x_given_z}
\vspace{-0.5em}
\end{eqnarray}
where $p({\bf x}; \theta({\bf z}))$ is a tractable density (e.g., Gaussian or Bernoulli) with the parameters $\theta({\bf z})$, the output of a deep model $\theta(\cdot)$ with its own weight parameters. In the unsupervised learning setup where we are given only ambient data $\{{\bf x}^n\}_{n=1}^N$, the model can be learned by the MLE, i.e., maximizing $\sum_{n=1}^N \log p({\bf x}^n)$. This requires posterior inference $p({\bf z}|{\bf x})$, but as the exact inference is intractable, the VAE adopts the variational technique: approximate $p({\bf z}|{\bf x}) \approx q({\bf z}|{\bf x})$, where $q({\bf z}|{\bf x})$ is a freely chosen tractable density 
\begin{equation}
q({\bf z}|{\bf x}) = q({\bf z}; \nu({\bf x})), 
\label{eq:q_z_given_x} 
\end{equation}
where $\nu({\bf x})$ is another deep model. A typical choice, assumed throughout the paper, is independent Gaussian, 
\begin{equation}
q({\bf z}|{\bf x}) 
  = \prod_{j=1}^d \mathcal{N}(z_j; m_j({\bf x}), s_j({\bf x})^2)
\label{eq:q_gaussian} 
\end{equation}
where $\nu({\bf x}) = \{ m_j({\bf x}), s_j({\bf x}) \}_{j=1}^d$ constitutes the mean and the variance parameters.

The data log-likelihood admits the ELBO as its lower bound, and we maximize it wrt both $\theta(\cdot)$ and $\nu(\cdot)$: 
\begin{equation}
\textrm{ELBO}(\theta,\nu) = 
  -\textrm{Recon}(\theta,\nu) - \mathbb{E}_{p_d({\bf x})} \big[ 
     \textrm{KL}( q({\bf z}|{\bf x}) || p({\bf z}) ) \big],
\label{eq:obj_vae}
\end{equation}
where $p_d({\bf x})$ is the empirical data distribution that represents our data $\{{\bf x}^n\}_{n=1}^N$, and 
\begin{equation}
\textrm{Recon}(\theta,\nu) = 
  -\mathbb{E}_{p_d({\bf x})} \big[ E_{q({\bf z}|{\bf x})} [ \log p({\bf x}|{\bf z}) ] \big]
\label{eq:recon_loss}
\end{equation}
is the reconstruction loss, identical to the negative expected log-likelihood.
Even though maximizing (\ref{eq:obj_vae}) can yield a model that explains the data well (i.e., high data likelihood), the learned model does not necessarily exhibit {\em disentanglement of latent factors}, as defined in the next section.

\subsection{Latent Disentanglement
}\label{sec:fvae}

We say that the latent vector ${\bf z}$ is disentangled if for each dimension $j=1,\dots,d$, varying $z_j$, while fixing other factors, results in the variation of the $j$-th aspect exclusively in the observation ${\bf x}$. For example, consider ${\bf x}$ to represent a face image, and let $j$ be the factor responsible for the facial pose (azimuth). Then varying $z_j$ while fixing other factors would generate images of different facial poses with other aspects, such as subject ID, gender, skin color, and facial expression, intact. 

To achieve this goal of disentanglement, the {\bf Factor-VAE}~\cite{factor_vae18} aims to minimize the following loss function\footnote{ 
There are several other VAE learning algorithms aiming for disentanglement in the similar flavor, and we briefly summarize the related work in Sec.~\ref{sec:related}.}: 
\begin{eqnarray}
\vspace{-0.5em}
\mathcal{L}_F &=& 
  \textrm{Recon}(\theta,\nu)
  + \mathbb{E}_{p_d({\bf x})} \Bigg[ \sum_{j=1}^d  \textrm{KL}( q(z_j|{\bf x}) || p(z_j) ) \Bigg] 
  \nonumber \\
&& \ \ \ \ + \ 
    \gamma \textrm{KL} \bigg( q({\bf z})|| \prod_{j=1}^d q(z_j)) \bigg).
\label{eq:obj_factor_vae}
\vspace{-0.5em}
\end{eqnarray}

In (\ref{eq:obj_factor_vae}) the first two terms correspond to the VAE, whereas the last term, known as the {\em total correlation} (TC)\footnote{
In the optimization, the difficult log-ratio between the mixtures (c.f.~(\ref{eq:q_z})) is circumvented by the density ratio estimation proxy~\cite{density_ratio_jordan,density_ratio_sugiyama}: they introduce and learn a discriminator $D({\bf z})$ that classifies samples from $q({\bf z})$ against those from $\prod_j q(z_j)$, and establish that $\log \frac{q({\bf z})}{\prod_j q(z_j)} \approx \log \frac{D({\bf z})}{1-D({\bf z})}$.
}, encourages factorization of the so-called {\em aggregate posterior} $q({\bf z})$, 
\begin{equation}
q({\bf z}) = \int q({\bf z}|{\bf x}) p_d({\bf x}) d{\bf x} 
    = \frac{1}{N} \sum_{n=1}^N q({\bf z}|{\bf x}^n).
\label{eq:q_z}
\end{equation}
That is, $q({\bf z})$ can be regarded as a {\em model-induced prior}, and the Factor-VAE imposes full independence of factors in this prior by penalizing $\textrm{KL}(q({\bf z})|| \prod_j q(z_j))$. An intuition is that as the TC encourages independence in the dimensions of ${\bf z}$, it also reduces the focus of $\mathcal{L}_F$ on the mutual information between ${\bf x}$ and ${\bf z}$ (the second term), leading to the model able to learn informative disentangled latent representations. 


Empirical results in~\cite{factor_vae18} have demonstrated that Factor-VAE is often able to achieve strong disentanglement performance.  However,  a key issue remains in that the model is unable to systematically discern meaningful latent factors from the nuisance ones because the approach relies on a heuristically chosen latent dimension $d$, sufficiently large to encompass all true relevant factors.  The lack of discrimination between relevant factors and nuisance in ${\bf z}$ may degrade the disentanglement performance and lead the model to learn redundant factors.  

In what follows, we address this issue by introducing relevance indicator variables that can be automatically learned from data together with the VAE parameters.


\section{Relevance Factor VAE (RF-VAE)}\label{sec:rfvae}

The key motivation of our approach is that for the factor $j$ to be relevant, its marginal model-induced prior $q(z_j)$ ought to be highly non-Gaussian, in contrast to the VAE's attempt, through the second term in (\ref{eq:obj_factor_vae}), to equally strongly match $q(z_j|{\bf x})$ to $p(z_j)=\mathcal{N}(0,1)$ across all ${\bf x}$.  This can be easily seen: $q(z_j)$ is a mixture of Gaussians (c.f., (\ref{eq:q_z}) and (\ref{eq:q_gaussian})) with components $q(z_j|{\bf x}) = \mathcal{N}(z_j; m_j({\bf x}), s_j({\bf x})^2)$.  If the factor $j$ is relevant,  $z_j$ should never be independent of ${\bf x}$, reinforcing the non-Gaussianity of the mixture as the whole, where each component differs from another. On the other hand, for a nuisance dimension $j'$, $z_{j'}$ are, by definition, independent of ${\bf x}$, i.e., $q(z_{j'}|{\bf x}) = q(z_{j'})$, enabling the second KL term in (\ref{eq:obj_factor_vae}) to vanish and have no effect on $\mathcal{L}_F$.

To differentiate the prior KL losses for relevant factors from those of the nuisance ones, we partition the latent dimensions into two disjoint subsets, ${\bf R}$ (relevant) and ${\bf N}$ (nuisance), i.e., $\{1,\dots,d\} = {\bf R} \cup {\bf N}$, ${\bf R} \cap {\bf N} = \emptyset$. Critically, KL penalties on ${\bf R}$ and ${\bf N}$ need to be distinct and learned from the data.  With that in mind, in Sec.~\ref{sec:rfvae_cheat} we first develop the model where we assume that ${\bf R}$ and ${\bf N}$ are known. We subsequently, in Sec.~\ref{sec:rfvae_no_cheat}, relax this constraint to, in a principled manner, learn the partition directly from data.

\subsection{RF-VAE: Known ${\bf R}$}\label{sec:rfvae_cheat}

Assuming that we know the index sets ${\bf R}$ and ${\bf N}$, we propose the following loss function for the disentangled VAE learning:
\begin{eqnarray}
\lefteqn{
\mathcal{L}_{R_0} \ = \ 
  \textrm{Recon}(\theta,\nu) 
   + \mathbb{E}_{p_d({\bf x})} \Bigg[ \sum_{j=1}^d  \lambda_j \textrm{KL}( q(z_j|{\bf x}) || p(z_j) ) \Bigg] } \nonumber \\ 
&& \ \ \ \ + \
  \gamma \textrm{KL} \bigg( q({\bf z}_{\bf R})|| \prod_{j\in{\bf R}} q(z_j)) \bigg),  
\label{eq:obj_rfvae_cheat} \\
&\textrm{where}& \lambda_j = 
  \left \{ \begin{array}{ll}
    \lambda_{min} 
    & \textrm{if $j\in{\bf R}$} \vspace{+0.5em} \\
    \lambda_{max} 
    & \textrm{if $j\in{\bf N}$}
  \end{array} \right. \ \ \ \ 
  (\lambda_{min} < \lambda_{max}),
\nonumber 
\end{eqnarray}
e.g., $\lambda_{min}=0.1$ and $\lambda_{max}=10.0$.

As shown, we have made two modifications from the loss function of Factor-VAE: i) The prior KL loss is penalized differently according to the relevance of each dimension $j$, penalizing less for $j\in{\bf R}$ with impact $\lambda_{min}$, and more for $j\in{\bf N}$ with impact $\lambda_{max}$. ii) The TC takes into account only the relevant dimensions.  
In the Supplement, we provide a theoretical justification for this approach. In particular, our analysis supplies a rigorous theoretical underpinning for why minimizing TC leads to factor disentanglement, beyond just the intuitive argument made previously in Factor-VAE and other related approaches. 

\noindent\textbf{Optimization}. To optimize (\ref{eq:obj_rfvae_cheat}), we follow the approach similar to that taken in Factor-VAE. The TC term is approximated by the density ratio proxy  
\begin{equation}
\textrm{KL} \bigg( q({\bf z}_{\bf R})|| \prod_{j\in{\bf R}} q(z_j) \bigg) \approx 
  \mathbb{E}_{q({\bf z}_{\bf R})} \bigg[ 
    \log \frac{D({\bf z}_{\bf R})}{1-D({\bf z}_{\bf R})} \bigg], 
\label{eq:rfvae_density_ratio} 
\end{equation}
where $D(\cdot)$ is the discriminator that discerns samples from $q({\bf z}_{\bf R})$ (as output $1$) from those in $\prod_{j\in{\bf R}} q(z_j)$ (as output $0$). That is, 
\begin{equation}
\max_{D} \Big( 
    \mathbb{E}_{{\bf z}\sim q({\bf z})}[\log D({\bf z}_{\bf R})] + 
    \mathbb{E}_{{\bf z}\sim \prod_j q(z_j)}[\log (1-D({\bf z}_{\bf R}))]
\Big).
\label{eq:rfvae_D}
\end{equation}
In the optimization, we alternate gradient updates for (\ref{eq:rfvae_D}) wrt $D(\cdot)$ and (\ref{eq:obj_rfvae_cheat}) wrt the VAE parameters with the TC term replaced by the expected log-ratio (\ref{eq:rfvae_density_ratio}).

\subsection{RF-VAE: Learning ${\bf R}$}\label{sec:rfvae_no_cheat}

Our previous assumption, that the index set of relevant dimensions ${\bf R}$ is known, is often times not practical. In this section we propose a principled way to learn the relevant dimensions automatically from data. 

The key idea is to introduce a {\em relevance vector} ${\bf r}$, of the same dimension as ${\bf z}$, where $r_j=1$ ($0$) indicates that $z_j$ is a relevant (resp., nuisance) factor, for $j=1,\dots,d$. 
We can learn ${\bf r}$ together with the VAE parameters 
by optimizing a loss function similar to $\mathcal{L}_{R_0}$ in (\ref{eq:obj_rfvae_cheat}). Specifically, since ${\bf r}$ defines the relevance set ${\bf R} = \{j: r_j=1\}$, we can formulate an optimization problem that minimizes $\mathcal{L}_{R_0}$ for a given ${\bf r}$, and regularizes ${\bf r}$ to discover a minimally redundant set of relevant factors. 

With the density ratio approximation for the TC term, we incorporate the optimization variables ${\bf r}$ in $\mathcal{L}_{R_0}$ in the following way:
\begin{eqnarray}
\vspace{-0.5em}
\lefteqn{
  \textrm{Recon}(\theta,\nu) 
   + \mathbb{E}_{p_d({\bf x})} \Bigg[ \sum_{j=1}^d  \lambda(r_j) \textrm{KL}( q(z_j|{\bf x}) || p(z_j) ) \Bigg] } \nonumber \\ 
&& \ + \
  \gamma \mathbb{E}_{q({\bf z})} \bigg[ 
    \log \frac{D({\bf r} \circ {\bf z})}{1-D({\bf r} \circ {\bf z})} \bigg] + 
  \eta_S ||{\bf r}||_1, \ \ \ \ \ \ \
\label{eq:obj_rfvae_nocheat_pre} 
\end{eqnarray}
where $\lambda(\cdot)$ is a decreasing function\footnote{In our experiments, we simply choose a linear function.} with $\lambda(0)=\lambda_{max} > \lambda(1)=\lambda_{min}$, and $\circ$ is the element-wise (Hadamard) product.  Note that the last L1 term penalizes too many dimensions to be chosen as relevant, encouraging minimal redundancy. The remaining difference from $\mathcal{L}_{R_0}$ is the TC term, where the discriminator $D(\cdot)$ now takes the latent vector scaled by ${\bf r}$ as its input.  For given ${\bf r}$, the discriminator is learned from the following optimization:
\begin{equation}
\vspace{-0.2em}
\max_{D} 
    \mathbb{E}_{{\bf z}\sim q({\bf z})}[\log D({\bf r} \circ {\bf z})] + 
    \mathbb{E}_{{\bf z}\sim \prod_j q(z_j)}[\log (1-D({\bf r} \circ {\bf z}))]
\label{eq:rfvae_nocheat_D}
\vspace{-0.3em}
\end{equation}

This allows one to mitigate the impact of nuisance dimensions ($r_j=0$) on the TC term while leaving the relevant latent variables $z_j$ ($r_j=1$) intact. 


Since optimizing (\ref{eq:obj_rfvae_nocheat_pre}) wrt ${\bf r} \in \{0,1\}^d$ is a difficult combinatorial problem, we relax ${\bf r}$ to be a continuous space ${\bf r} \in [0,1]^d$. Furthermore, to encourage each $r_j$ to be close to either $0$ or $1$ and discourage fractional values, 
we include the entropic loss, $H({\bf r}) = -\sum_{j=1}^d \big( r_j \log r_j + (1-r_j) \log (1-r_j) \big)$. Our comprehensive loss function hence becomes:
\begin{eqnarray}
\lefteqn{
\mathcal{L}_{R}({\bf r}, \{\theta,\nu\}) \ = \ 
  \textrm{Recon}(\theta,\nu) \ + \
  \eta_S ||{\bf r}||_1 \ + \ \eta_H H({\bf r}) } \nonumber \\
&& \ \ \ \ + \ \ 
   \mathbb{E}_{p_d({\bf x})} \Bigg[ \sum_{j=1}^d  \lambda(r_j) \textrm{KL}( q(z_j|{\bf x}) || p(z_j) ) \Bigg] \nonumber \\
&& \ \ \ \ +  \ \ \gamma \mathbb{E}_{q({\bf z})} \bigg[ 
    \log \frac{D({\bf r} \circ {\bf z})}{1-D({\bf r} \circ {\bf z})} \bigg] 
    \ \ \ \ \ \ \ \ \ \ \ \ \ \ \  \ \ \ \ \ \ \ \ \ \ \ \ 
 \label{eq:obj_rfvae_nocheat} 
\end{eqnarray}
Again, this loss is minimized by alternating the gradient updates for (\ref{eq:rfvae_nocheat_D}) wrt $D(\cdot)$ and (\ref{eq:obj_rfvae_nocheat}) wrt both the VAE parameters $\{\theta,\nu\}$ and the relevance vector ${\bf r}$.

\section{Related Work
}\label{sec:related}

Most approaches to latent disentanglement consider the learning objectives combining the ELBO loss in (\ref{eq:obj_vae}) with the regularization terms that encourage prior latent factor independence. We summarize some key recent approaches below.
\begin{itemize}
\item {\bf $\beta$-VAE}~\cite{beta_vae17}. 
Instead of directly introducing $\textrm{KL}(q({\bf z})||p({\bf z}))$, the challenge of dealing with non-factorized (\ref{eq:q_z}) is circumvented through adoption of the {\em averaged} divergence between the posterior and the prior. Hence, the objective\footnote{As the penalty term coincides with the KL term in the ELBO, one can merge the two while having $\beta\geq 1$.} to minimize is
\begin{equation}
\mathcal{L}(\theta,\nu) = -\textrm{ELBO} + \beta \frac{1}{N} \sum_{i=1}^N \textrm{KL}( q({\bf z}|{\bf x}^i) || p({\bf z}) ), 
\label{eq:obj_beta_vae}
\end{equation}
with $\beta \geq 0$ as the balancing constant.
\item {\bf AAE}~\cite{aae16}. The adopted regularization term is $\textrm{KL}(q({\bf z})||p({\bf z}))$. However, because of the difficulty of dealing with $q({\bf z})$ in the optimization process, the authors employ an adversarial learning strategy by introducing a discriminator $D({\bf z})$ that is adversarially learned to discriminate samples from $p({\bf z})$ against those from the non-factorized $q({\bf z})$.
%
\item {\bf Factor-VAE}~\cite{factor_vae18}. The full independence of $q({\bf z})$ is imposed by penalizing $\textrm{KL}(q({\bf z})|| \prod_j q(z_j))$ in the TC manner. Unlike AAE's implicit regularization, they do incorporate the KL term explicitly in the objective. The difficult-to-estimate log-ratio between the mixtures (c.f.~(\ref{eq:q_z})) is proxied by the density ratio estimation: for a near-optimal discriminator $D({\bf z})$ that classifies samples from $q({\bf z})$ against those from $\prod_j q(z_j)$, one can establish that $\log \frac{q({\bf z})}{\prod_j q(z_j)} \approx \log \frac{D({\bf z})}{1-D({\bf z})}$. Given $D$, the objective function to minimize is:
\begin{equation}
\mathcal{L}(\theta,\nu) = -\textrm{ELBO} + \beta \mathbb{E}_{q({\bf z})} \bigg[ \log \frac{D({\bf z})}{1-D({\bf z})} \bigg].
\label{eq:obj_factor_vae_2}
\end{equation}
In practice, every SGD update for (\ref{eq:obj_factor_vae}) is accompanied by the update of $D$ for the adversarial learning. 
\item \textbf{Adversarial Nonlinear ICA}~\cite{nica17} aims to learn the deterministic auto-encoder by minimizing the reconstruction error and the Jensen-Shannon divergence between $q({\bf z})$ and $\prod_j q(z_j)$, where the latter is optimized by adversarial learning. 
\item {\bf Other variants}. Other variants of the above approaches largely follow the same principles. For instance, DIP-VAE~\cite{dip_vae18} penalizes the deviation of $\mathbb{V}(q({\bf z}))$ from the identity, while InfoGAN~\cite{infogan16} aims to minimize the reconstruction error in the ${\bf z}$-space in addition to the conventional reconstruction error in the ${\bf x}$-space. The $\beta$-TCVAE algorithm~\cite{tcvae} aimed to derive a decomposable TC term, instead arriving at the same objective as the Factor-VAE. However, they employ the mini-batch weighted sampling strategy instead of the auxiliary adversarial discriminator training.
\end{itemize}

Some recent approaches seek to partition the latent variables into meaningful groups to achieve improved disentanglement. Since the factors underlying the data can oftentimes be a mix of categorical and real-valued sources, a hybrid model that jointly represents discrete and continuous latent variables was introduced in~\cite{hybrid_z}. In~\cite{lecun_disent}, they considered the setup where the data is partially labeled with specific class categories. In such setup, they separately treat the factors associated with the labels from those that are not, leading to an interesting conditional factor model. These approaches are related, but inherently different from our model in that we aim to simultaneously differentiate and identify the relevant latent variables from the nuisance factors. 



\section{Evaluation}\label{sec:evaluation}

In this section we evaluate our approaches on several benchmark datasets. The goodness of the disentanglement can be assessed both quantitatively and qualitatively, where the former requires the dataset to be fully factor-labeled; however, the labeled factors are only used for evaluation and not for model learning. Selecting appropriate quantitative scores is a key step in the evaluation process.  We consider a comprehensive suite of three metrics: i) the disentanglement metric proposed 
in~\cite{factor_vae18}, ii) our new disentanglement metric, and iii) the metrics studied in~\cite{williams18}. All metrics are briefly described in the following section.  Qualitative assessment is typically accomplished through visualizations of data synthesis via latent space traversal.  We include both in our experiments.

Specific to our models that separate relevant from nuisance variables is another evaluation metric, proposed in~\cite{lecun_disent}.  However, since this metric is not applicable to competing models with no stratification ability, we report its results in the Supplement.

\subsection{Disentanglement Metrics}\label{sec:disent_metrics}

\textbf{Metric I~\cite{factor_vae18} - One factor fixed}. The goal of this metric is to assess the variability of the discovered latent factors as a function of the true factor variance.  Let ${\bf v}$ be the vector of ground-truth factors for ${\bf x}$. For each factor index $j$, a set of $L$ samples ${\bf v}^{(i)} = (v_j, v^{(i)}_{-j})$, $i=1,\dots,L$, is collected, corresponding to clamping of factor $j$ (no variance) and free variation of the remaining factors. The existence of latent factor(s) $z_u$ with similar, vanishing variation would then indicate the discovery of known factor $j$.  In practice, this is accomplished by evaluating the encoder's outputs, ${\bf z}^{(i)} \sim q({\bf z}|{\bf x}^{(i)})$ for images ${\bf x}^{(i)}$ corresponding to samples ${\bf v}^{(i)}$. The (sample) variance ${\mathbb{V}}({\bf z})$ is used to find the index $u$ of the factor with the smallest variance, 
\begin{equation}
\vspace{-0.5em}
u := \arg\min_{1\leq j\leq d} {\mathbb{V}}(z_j).
\label{eq:metric_d}
\vspace{-0.0em}
\end{equation}
$u$ then serves as the covariate for predicting the true factor index $j$: the metric is defined to be the accuracy of a simple classifier that predicts $j$ from $u$, among all $j=1,\ldots,d$. If a model achieves strong disentanglement, we can expect $u = j$ (up to a permutation), making the classification easy. In particular, since $u,j \in \{1,\dots,d\}$, the data pairs $\{(u,j)\}$ can be represented as a simple contingency table, in which a majority vote classifier 
is used for prediction. We use $L=100$ samples to form each pair $(u,j)$, and collect $800$ pairs to compute the accuracies of the majority vote classifiers. 
Since the metric is 
based on random samples, 
we repeat the evaluation 
ten times randomly to report the means and standard deviations.

\textbf{Metric II (Our new metric) - One factor varied}. Following the notion of disentanglement in~\autoref{sec:fvae}, another reasonable approach is to collect samples with {\em only one factor varied}, instead of one factor fixed as in Metric I. That is, we collect images with ${\bf v}^{(i)} = (v^{(i)}_j, v_{-j})$ for $i=1,\dots,L$.  (\ref{eq:metric_d}) is then modified to $\arg\max$, and we can use the same majority vote classification to report the accuracy. Our evaluation results in the following sections demonstrate that this new metric shows higher agreement with qualitative assessment of disentanglement than Metric I. However, note that to compute Metric II the dataset needs to contain {\em dense} joint variations in all true factors, typically a reasonable assumption for large, diverse datasets. 

\textbf{Metric III~\cite{williams18}} proposed three metrics: 1) {\em Disentanglement}, 2) {\em Completeness}, and 3) {\em Informativeness}. These scores are regression-prediction based, using the latent vector ${\bf z}$ as the covariate for individual ground-truth factors $v_j$. Specifically, D measures the degree of dedication of each latent variable $z_k$ in predicting $v_j$ against others $v_{-j}$ (the higher, the better), C captures the degree of exclusive contribution of $z_k$ in predicting $v_j$ against others $z_{-k}$ (the higher, the better), and I measures the prediction error (the smaller, the better). For the regressors, both LASSO and Random Forests are used.

\subsection{Datasets and Results}\label{sec:datasets}

\textbf{Datasets summary}. We test all methods 
on the following datasets: \texttt{3D-Faces}~\cite{3dfaces}, \texttt{Sprites}~\cite{dsprites}, and \texttt{Teapots}~\cite{williams18}.  Results for \texttt{Celeb-A}~\cite{celeba} are provided in the Supplement.  All datasets provide ground-truth factor labels. 
For all datasets, the image sizes are normalized to $64 \times 64$, and the pixel intensity/color values are scaled to $[0,1]$.  We use cross entropy loss as the reconstruction loss.

\textbf{Model architectures}. We adopt the model architectures similar to those in~\cite{factor_vae18}. The encoders consist of 5-layer conv-nets followed by two fully connected layers, and the decoders are 4-layer deconv-nets after two fully connected layers. We apply $(4 \times 4)$ filters for the convolution and the transposed convolution (deconv) in both models. For the adversarial discriminator $D$ used for optimizing the TC loss, we use a 6-layer MLP model with 1000 hidden units per layer and the leaky ReLU nonlinearity. 
See Supplement for more details.

\textbf{Optimization parameters}. The optimization parameters are chosen similarly as those in~\cite{factor_vae18}. We use Adam with the batch size $64$. We run $3\times 10^5$ batch iterations. 
The trade-off parameters (i.e., $\beta$ for $\beta$-VAE, $\gamma$ for Factor-VAE and RF-VAE models, and  $(\eta, \lambda)$ for our RF-VAE models) are obtained through grid search to yield the best performance. 

Our two models are marked as: \textbf{RF-VAE-0}, which exploits the ground-truth ${\bf R}$  (Sec.~\ref{sec:rfvae_cheat}) and \textbf{RF-VAE}, which learns ${\bf R}$ from data  (Sec.~\ref{sec:rfvae_no_cheat}). In RF-VAE, the relevance vector ${\bf r}$ is initialized as all-$0.5$ vector. We contrast our models to \textbf{Vanilla VAE}~\cite{vae14}, \textbf{$\beta$-VAE}~\cite{beta_vae17}, and \textbf{Factor-VAE}~\cite{factor_vae18}.

\subsubsection{3D-Faces}\label{sec:3d_faces}

\begin{figure}
\begin{center}
\includegraphics[trim = 8mm 0mm 10mm 4mm, clip, scale=0.315]{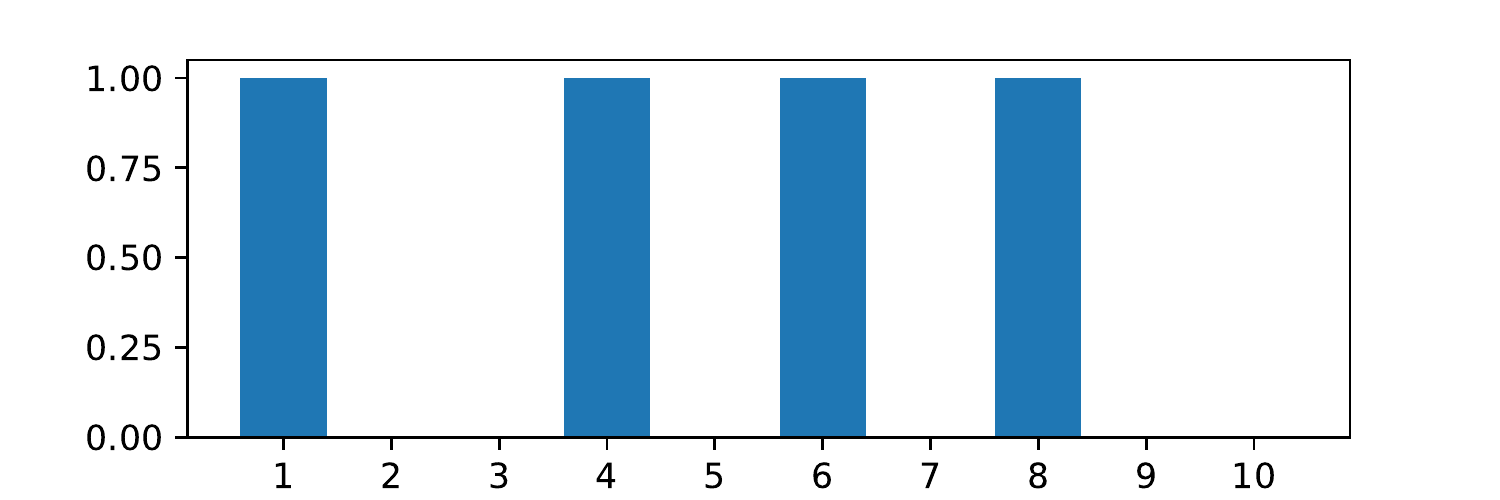} 
\includegraphics[trim = 14mm 0mm 15mm 4mm, clip, scale=0.315]{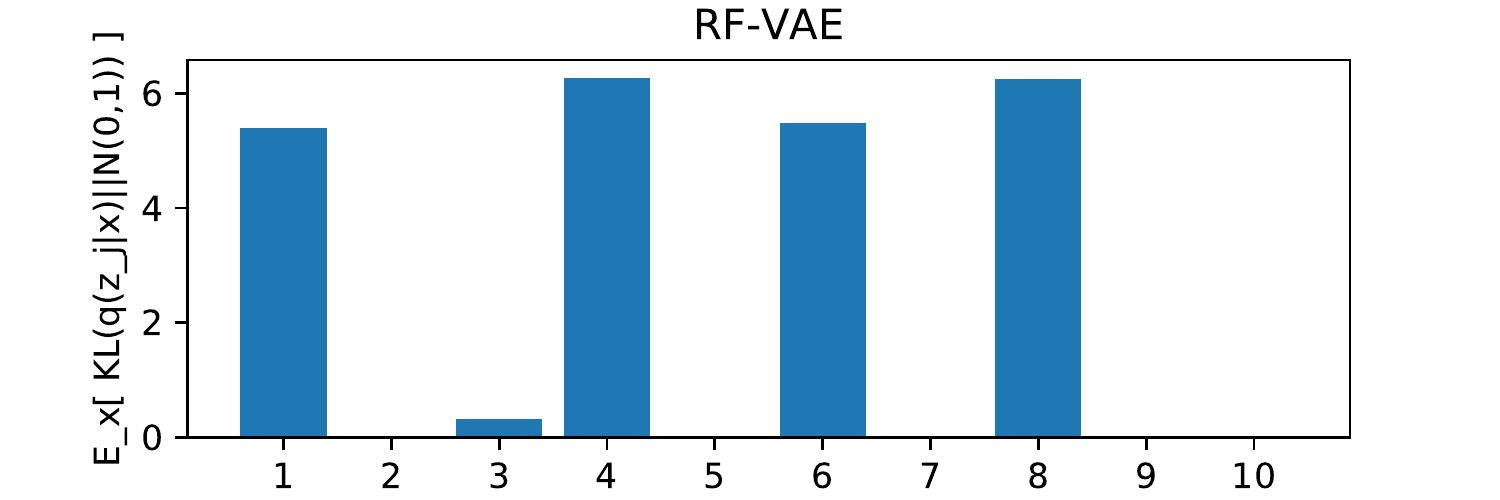}
\end{center}
\vspace{-1.3em}
\caption{RF-VAE on 3D-Faces dataset. (Left) Learned relevance ${\bf r}$. (Right) Expected prior KL divergences for individual dimensions. 
}
\label{fig:rfvae_learned_3dfaces}
\vspace{-1.5em}
\end{figure}

This dataset contains synthetic face images rendered from a 3D face model~\cite{3dfaces}, with four factors of variation (number of modes): subject ID (50), azimuth (21), elevation (11), and lighting along the horizon (11). 

We choose the latent dimension $d=10$ for all competing models to encompass the true $4$. 
Quantitative results are shown in ~\autoref{tab:3dfaces_metrics}. All models attain perfect Metric I scores. For other metrics, RF-VAE-0 (with known ${\bf R}=\{1,2,3,4\}$) consistently outperforms other models. Interestingly, RF-VAE, which learns the relevance set ${\bf R}$ from data, performs nearly equally well as RF-VAE-0, and significantly better than Factor-VAE in terms of Metric II.


Result in ~\autoref{tab:3dfaces_metrics} suggest that Metric I score may only be partially indicative of the underlying disentanglement performance: all models attain a perfect Metric I scores. Differentiation is more prominent, however, in Metrics II and III.  This is not surprising because it is possible to achieve a high Metric I score when the same factor is shared across multiple latent variables, i.e., it may suffer from redundancy in representation. However, redundancy would decrease Completeness. On the other hand, our Metric II can simultaneously capture both D and C.


\begin{table*}[t]
\centering
\caption{Disentanglement metrics for benchmark datasets.  For Metric III, the three figures in each cell indicate Disentanglement / Completeness / Informativeness (top row based on the LASSO regressor, the bottom on the Random Forest. Note that the higher the better for D and C, while the lower the better for I. The best scores for each metric (within the margin of significance) among the competing models are shown in red and second-best in blue.}

\begin{tabular}{c}

\begin{subtable}[tc]{\textwidth}
\caption{Disentanglement metrics on the \texttt{3D-Faces} dataset. 
}
\label{tab:3dfaces_metrics}

\begin{small}
\begin{sc}
\centering
\begin{tabular}{lccccc}
\toprule
 & Vanilla-VAE & $\beta$-VAE & Factor-VAE & RF-VAE-0 & RF-VAE \\
\midrule
Metric I & ${\color[rgb]{1,0,0} 100.0 \pm 0.00}$ & ${\color[rgb]{1,0,0} 100.0 \pm 0.00}$ & ${\color[rgb]{1,0,0} 100.0 \pm 0.00}$ & ${\color[rgb]{1,0,0} 99.80 \pm 0.14}$ & ${\color[rgb]{1,0,0} 99.93 \pm 0.06}$ \\ \hline
Metric II & $93.44 \pm 0.74$ & ${\color[rgb]{0,0,1} 95.48 \pm 0.62}$ & $92.78 \pm 1.09$ & ${\color[rgb]{1,0,0} 99.61 \pm 0.26}$ & ${\color[rgb]{0,0,1}95.19} \pm 0.53$ \\ \hline
\multirow{2}{*}{Metric III} 
& {\color[rgb]{0,0,1}0.96} / 0.81 / {\color[rgb]{0,0,1}0.37} & {\color[rgb]{0,0,1}0.96} / 0.78 / 0.40 & {\color[rgb]{1,0,0} {1.00}} / 0.82 / {\color[rgb]{1,0,0}{0.36}} & {\color[rgb]{1,0,0}{1.00}} / {\color[rgb]{0,0,1}0.97} / 0.51 & {\color[rgb]{1,0,0}{1.00}} / {\color[rgb]{1,0,0}{1.00}} / 0.48 \\ 
& {\color[rgb]{0,0,1}0.99} / 0.84 / {\color[rgb]{0,0,1}0.26} & 0.98 / 0.86 / 0.31 & 0.96 / 0.83 / {\color[rgb]{1,0,0}{0.25}} & {\color[rgb]{1,0,0}{1.00}} / {\color[rgb]{1,0,0}{0.95}} / 0.46 &  {\color[rgb]{1,0,0}{1.00}} / {\color[rgb]{0,0,1}0.93} / 0.37 \\
\bottomrule
\end{tabular}
\end{sc}
\end{small}
\end{subtable}

\\

\begin{subtable}[t]{\textwidth}
\caption{Disentanglement metrics on the \texttt{Sprites} dataset.}
\label{tab:dsprites_metrics}
\begin{small}
\begin{sc}
\begin{tabular}{lccccc}
\toprule
 & Vanilla-VAE & $\beta$-VAE & Factor-VAE & RF-VAE-0 & RF-VAE \\
\midrule
Metric I & $80.20 \pm 0.33$ & $80.81 \pm 0.78$ & $81.98 \pm 1.01$ & ${\color[rgb]{1,0,0} 91.99 \pm 0.85}$ & ${\color[rgb]{0,0,1} 85.35 \pm 1.16}$ \\ \hline
Metric II & $58.15 \pm 1.44$ & $76.83 \pm 0.89$ & $77.64 \pm 1.42$ & ${\color[rgb]{1,0,0}  85.41 \pm 1.48}$ & ${\color[rgb]{0,0,1}79.08 \pm 1.31}$ \\ \hline
\multirow{2}{*}{Metric III} 
& 0.59 / 0.68 / {\color[rgb]{1,0,0} 0.52} & 0.67 / 0.69 / {\color[rgb]{0,0,1}0.53} & 0.84 / 0.84 / {\color[rgb]{0,0,1}0.53} & {\color[rgb]{1,0,0} 0.89} / {\color[rgb]{1,0,0}1.00} / 0.64 & {\color[rgb]{0,0,1}0.85 }/ {\color[rgb]{0,0,1}0.87} / {\color[rgb]{0,0,1}0.53} \\ 
& 0.57 / 0.69 / 0.46 & 0.72 / {\color[rgb]{0,0,1}0.84} / {\color[rgb]{1,0,0} 0.40} & {\color[rgb]{0,0,1}0.73} / 0.82 / {\color[rgb]{0,0,1}0.41} & {\color[rgb]{1,0,0} 0.78} / {\color[rgb]{1,0,0} 0.94} / 0.62 & {\color[rgb]{0,0,1}0.73} / 0.83 / {\color[rgb]{0,0,1}0.41} \\
\bottomrule
\end{tabular}
\end{sc}
\end{small}
\end{subtable}

\\

\begin{subtable}[t]{\textwidth}
\caption{Disentanglement metrics on the \texttt{Sprites Oval-Only} dataset. }
\label{tab:dsprites_oval_metrics}
\begin{small}
\begin{sc}
\begin{tabular}{lccccc}
\toprule
 & Vanilla-VAE & $\beta$-VAE & Factor-VAE & RF-VAE-0 & RF-VAE \\
\midrule
Metric I & ${\color[rgb]{0,0,1}97.19 \pm 0.42}$ & $75.33 \pm 0.64$ & ${\color[rgb]{1,0,0} 100.0 \pm 0.00}$ & ${\color[rgb]{1,0,0} 100.0 \pm 0.00}$ & ${\color[rgb]{1,0,0} 100.0 \pm 0.00}$ \\ \hline
Metric II & $53.23 \pm 1.47$ & $70.20 \pm 1.19$ & ${\color[rgb]{0,0,1}80.59 \pm 1.05}$ & ${\color[rgb]{1,0,0} 95.96 \pm 0.44}$ & ${\color[rgb]{1,0,0} 95.40 \pm 0.47}$ \\ \hline
\multirow{2}{*}{Metric III} 
        & 0.42 / 0.43 / 0.54 & 0.58 / 0.49 / {\color[rgb]{0,0,1}0.49} & {\color[rgb]{1,0,0}1.00} / 0.88 / {\color[rgb]{1,0,0}0.33} & {\color[rgb]{0,0,1}0.97} / {\color[rgb]{0,0,1}0.93} / {\color[rgb]{1,0,0}0.33} & {\color[rgb]{1,0,0}1.00} / {\color[rgb]{1,0,0}0.99} / {\color[rgb]{0,0,1}0.49} \\ 
& 0.32 / 0.55 / 0.46 & 0.56 / 0.58 / 0.36 & 0.81 / 0.84 / {\color[rgb]{0,0,1}0.24} & {\color[rgb]{1,0,0}0.97} / {\color[rgb]{1,0,0}0.96} / 0.29 & {\color[rgb]{0,0,1}0.93} / {\color[rgb]{0,0,1}0.87} / {\color[rgb]{1,0,0}0.22} \\
\bottomrule
\end{tabular}
\end{sc}
\end{small}
\end{subtable}

\\

\begin{subtable}[t]{\textwidth}
\caption{Disentanglement metrics on the \texttt{Teapots} dataset.}
\label{tab:teapots_metrics}
\begin{small}
\begin{sc}
\begin{tabular}{lccccc}
\toprule
 & Vanilla-VAE & $\beta$-VAE & Factor-VAE & RF-VAE-0 & RF-VAE \\
\midrule
Metric I & $90.14 \pm 0.90$ & $56.94 \pm 1.14$ & ${\color[rgb]{0,0,1}91.93 \pm 0.83}$ & $78.00 \pm 1.02$ & ${\color[rgb]{1,0,0} 98.68 \pm 0.35}$ \\ \hline
Metric II & ${\color[rgb]{0,0,1}77.74 \pm 1.26}$ & $47.30 \pm 0.85$ & $74.61 \pm 1.75$ & $60.30 \pm 1.92$ & ${\color[rgb]{1,0,0} 83.10 \pm 1.22}$ \\ \hline
\multirow{2}{*}{Metric III} 
& {\color[rgb]{0,0,1}0.60} / 0.53 / {\color[rgb]{0,0,1}0.40} & 0.31 / 0.27 / 0.72 & {\color[rgb]{1,0,0}0.63} / {\color[rgb]{1,0,0}0.61} / 0.46 & 0.40 / 0.38 / 0.57 & {\color[rgb]{1,0,0}0.63} / {\color[rgb]{0,0,1}0.56} / {\color[rgb]{1,0,0}0.37} \\ 
& {\color[rgb]{0,0,1}0.81} / 0.72 / 0.31 & 0.45 / 0.61 / 0.52 & 0.75 / {\color[rgb]{0,0,1}0.78} / {\color[rgb]{0,0,1}0.29} & 0.62 / 0.59 / 0.49 & {\color[rgb]{1,0,0}0.90} / {\color[rgb]{1,0,0}0.79} / {\color[rgb]{1,0,0}0.27} \\
\bottomrule
\end{tabular}
\end{sc}
\end{small}
\end{subtable}

\end{tabular}

\end{table*}

For our RF-VAE, we also depict the learned relevance vector ${\bf r}$ in ~\autoref{fig:rfvae_learned_3dfaces}. It shows that our model discovers four factors, (1,4,6,8), signifying its ability to identify the true source of variation from the data alone. The found latent dimensions have higher (non-zero) expected prior KL divergences, $\mathbb{E}_{p_d({\bf x})} \big[ \textrm{KL}( q(z_j|{\bf x}) || p(z_j) ) \big]$, compared those in the nuisance factors (zero), as shown on the right panel of ~\autoref{fig:rfvae_learned_3dfaces}. As designed and anticipated, our model successfully tolerates large prior KL divergence matching penalty. 
On the other hand, the number of latent dimensions with large KL divergences (greater than $0.5$) for other models do not match the true number of factors; they are $8$ (Vanilla-VAE), $6$ ($\beta$-VAE), and $7$ (Factor-VAE). This suggests those models learn redundant or non-exclusive latent representations to explain the variation in the data, a property that can diminish disentanglement.  This may be a main reason why the competing models underperform RF-VAE. 

\begin{figure*}[t!]
\centering

\begin{subfigure}[b]{0.3\textwidth}
\includegraphics[trim = 8mm 2mm 8mm 0mm, clip, scale=0.3]{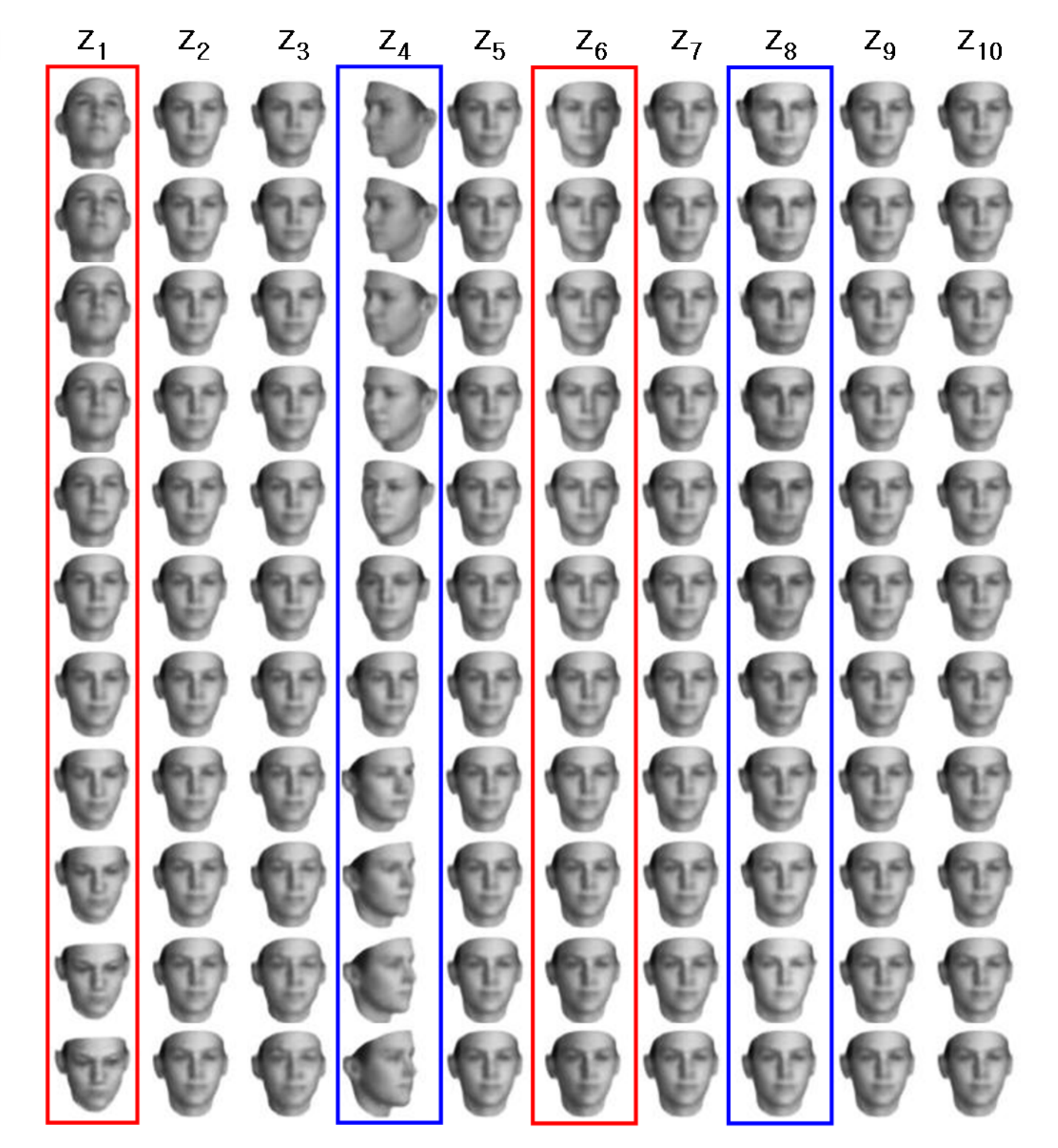}
\caption{
\texttt{3D-Faces} dataset. 
}
\label{fig:rfvae_traverse_3dfaces}
\end{subfigure}
~
\begin{subfigure}[b]{0.3\textwidth}
\includegraphics[trim = 9mm 2mm 7mm 0mm, clip, scale=0.3]{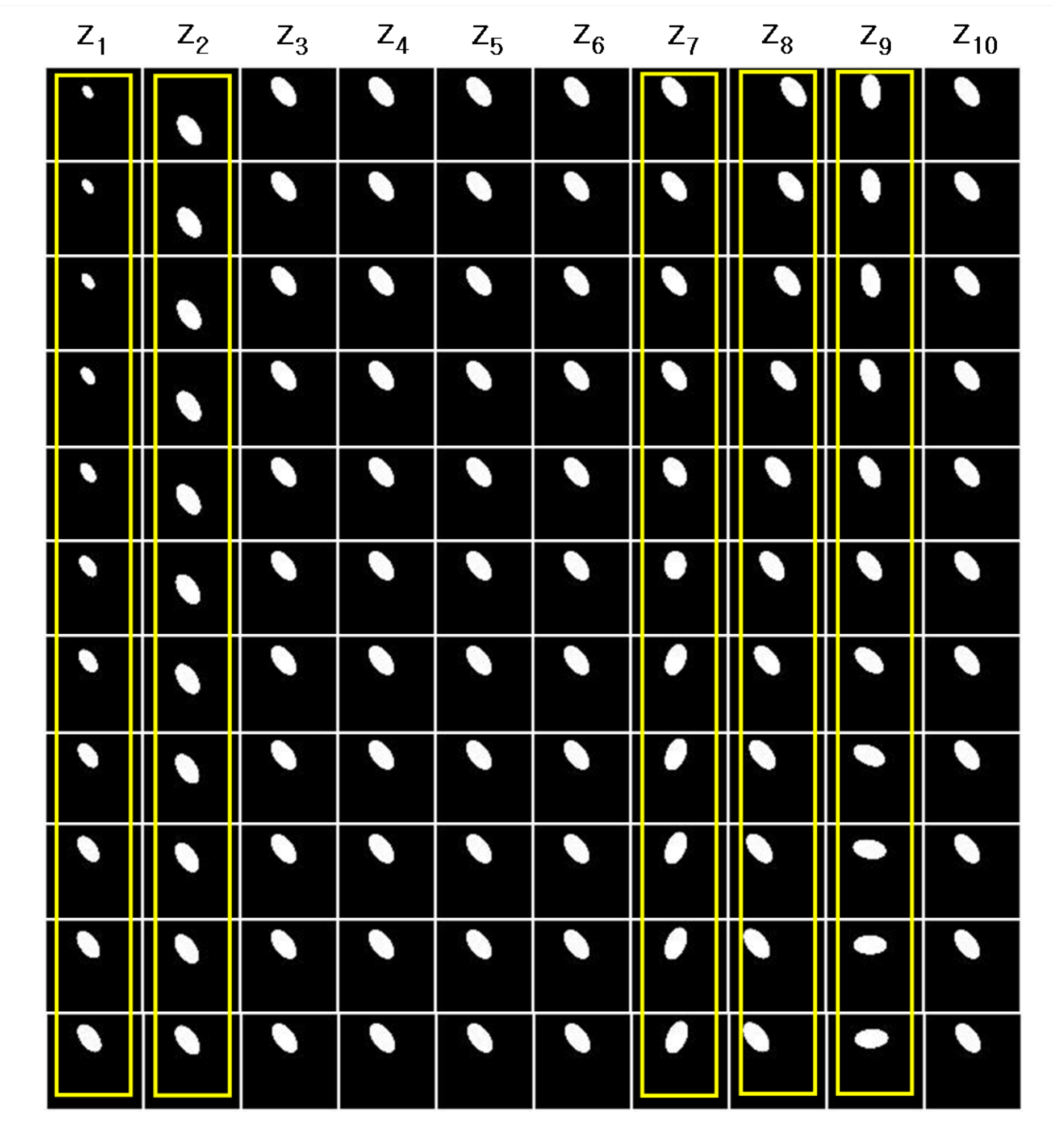}
\caption{
\texttt{Sprites-Oval-Only} dataset. 
}
\label{fig:rfvae_traverse_dsprites_oval}
\end{subfigure}
~
\begin{subfigure}[b]{0.3\textwidth}
\includegraphics[trim = 9mm 2mm 7mm 0mm, clip, scale=0.3]{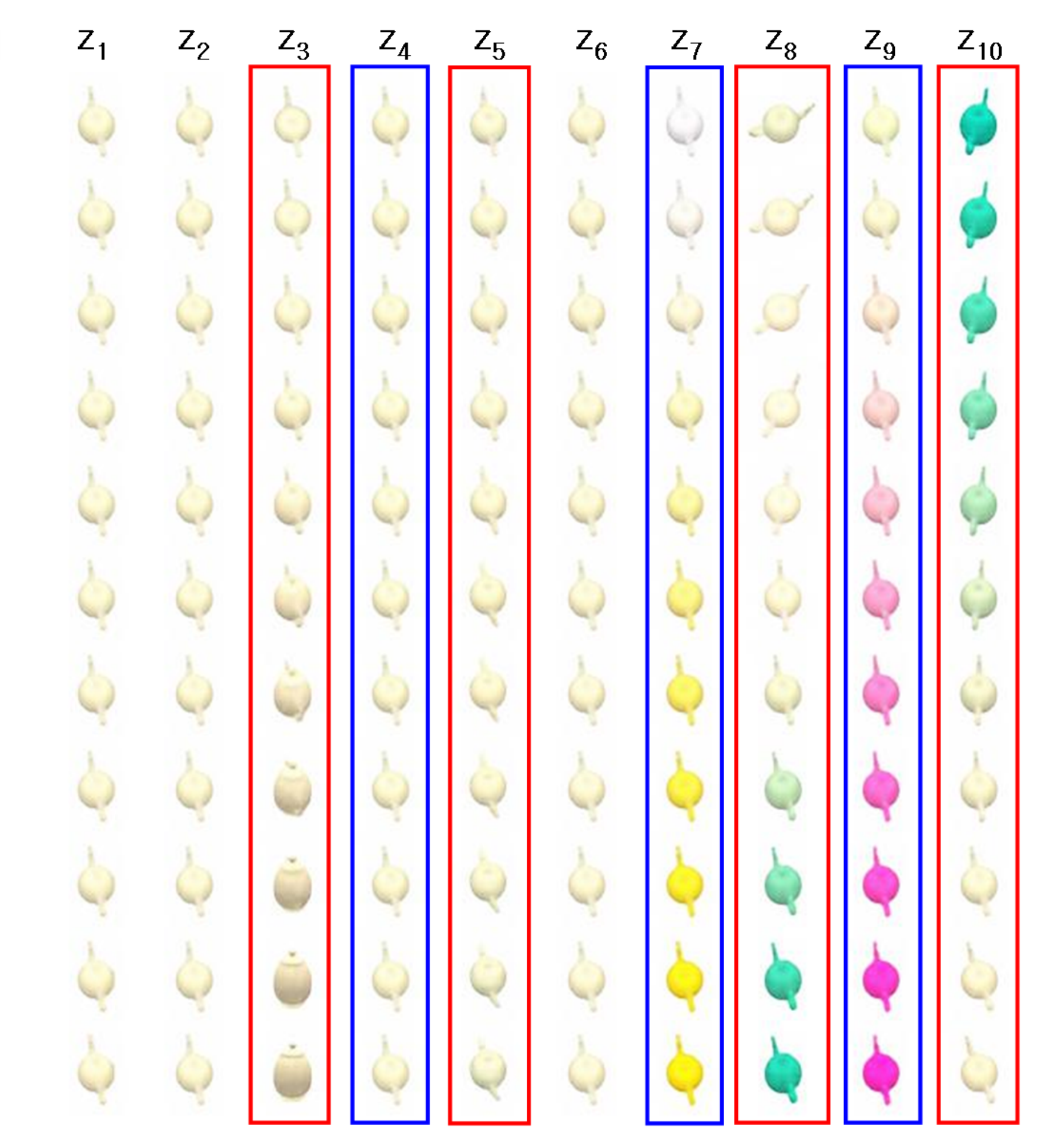}
\caption{
\texttt{Teapots} dataset. 
}
\label{fig:rfvae_traverse_teapots}
\end{subfigure}

\vspace{-1.0em}
\caption{Latent space traversals in RF-VAE on \texttt{3D-Faces}, \texttt{Sprites}, and \texttt{Teapots} datasets. 
(a) The four factors ($z_1$, $z_4$, $z_6$, $z_8$) recovered by our RF-VAE are highlighted within colored boxes. Variation in each of the four factors yields changes in a single aspect of synthesized images ($z_1=$ elevation, $z_4=$ azimuth, $z_6=$ lighting, $z_8=$ subject ID). 
(b) The recovered, highlighted, factors are ($z_1$, $z_2$, $z_7$, $z_8$, $z_9$). $z_1$ corresponds to scale, $z_2$ $Y$-pos, $z_8$ explains $X$-pos, and both $z_7$ and $z_{9}$ captures rotation.  
In all cases, other (non-highlighted) nuisance dimensions have little impact on image generation.
(c) The seven recovered factors ($z_3$, $z_4$, $z_5$, $z_7$, $z_8$, $z_9$, $z_{10}$) represent: $z_9$ the R channel, both $z_8$ and $z_{10}$ have the G channel and azimuth entangled in each. $z_7$ explains yellow color variation meaning (R,G) are entangled in it. Both $z_4$ and $z_5$ are numb, and $z_3$ is responsible for elevation, but  color variations are slightly entangled together.
}
\vspace{-1.0em}
\end{figure*}

Latent traversal results for RF-VAE are shown in ~\autoref{fig:rfvae_traverse_3dfaces}, where we depict images synthesized by the traversal of a single latent variable at a time. Visually, each relevant latent variable is strongly associated with one of the four factors, while the variation of nuisance variables has little impact on the generated images.

\subsubsection{Sprites 
}\label{sec:dsprites}

The dataset 
consists of $737,280$ binary 
images of three shapes (oval, square, and heart), undergoing variations in four geometric factors: scale (6 variation modes), rotation (40), and $X$, $Y$ translation (32 modes each), resulting in five factors total. 

\begin{figure}
\begin{center}
\includegraphics[trim = 8mm 0mm 10mm 4mm, clip, scale=0.315]{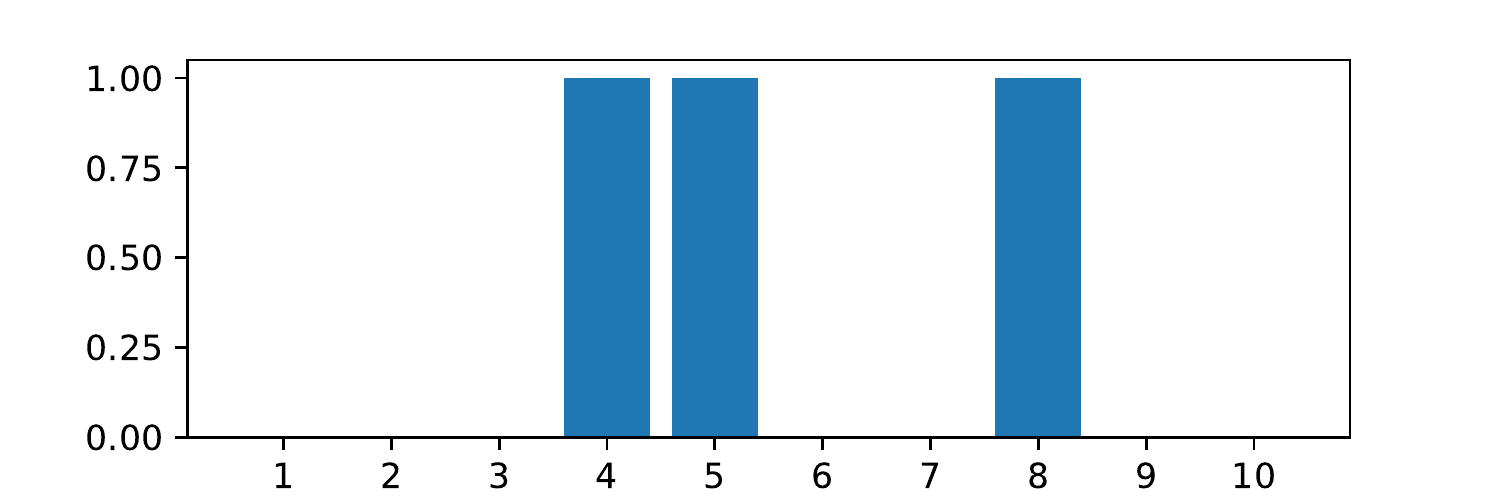} 
\includegraphics[trim = 14mm 0mm 15mm 4mm, clip, scale=0.315]{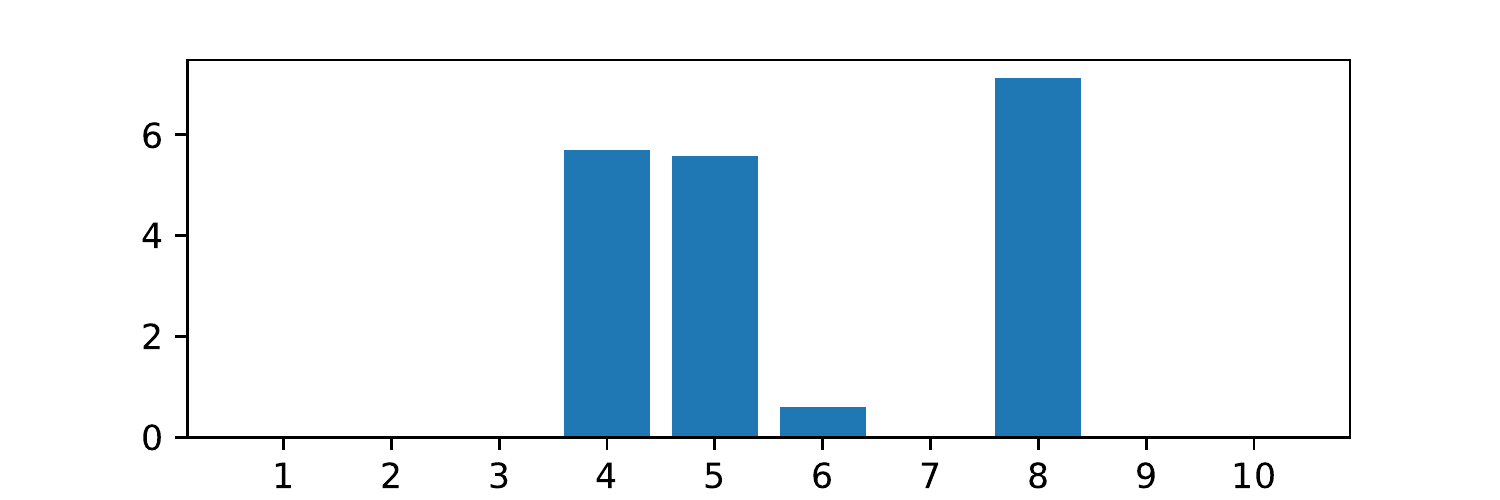}
\end{center}
\vspace{-1.5em}
\caption{RF-VAE on \texttt{Sprites}. (Left) Learned relevance vector ${\bf r}$. (Right) Expected prior KLs for individual dimensions. 
}
\label{fig:rfvae_learned_original_dsprites}
\end{figure}

For all competing models, we choose the latent dimension $d=10 > 5$. The disentanglement scores are reported in~\autoref{tab:dsprites_metrics}. Although RF-VAE-0, by exploiting the known number of factors, largely attained the highest scores, the scores are not perfect (e.g., Metric II far below $100\%$). 
As shown in the latent traversal results in the Supplement, 
the shape factor remains entangled with other factors in all latent variables $z_{\bf R}$.  RF-VAE similarly failed to identify the five relevant factors, elucidating three modes of variation (\autoref{fig:rfvae_learned_original_dsprites}), even though it performs slightly better than Factor-VAE.

One reason for this failure may lie in the difficulty of representing the shape factor, which is discrete in nature and has low mode cardinality\footnote{Note that while other factors also possess discrete modes, their cardinality is higher than that of the shape.};  hence, it may not be properly modeled by the continuous latents in the VAE. A solution may be to model hybrid latents c.f.,~\cite{hybrid_z}. Instead, we conduct experiments by eliminating the discrete shape factor: we consider a subset of oval-only images and retain the remaining geometric sources of variation. 





\noindent\textbf{Oval Shape Subset}. From the original dataset, we select the oval shape images, yielding four independent factors of variation across $245,760$ images.

\begin{figure}
\begin{center}
\includegraphics[trim = 9.87mm 0mm 8.7mm 5mm, clip, scale=0.313]{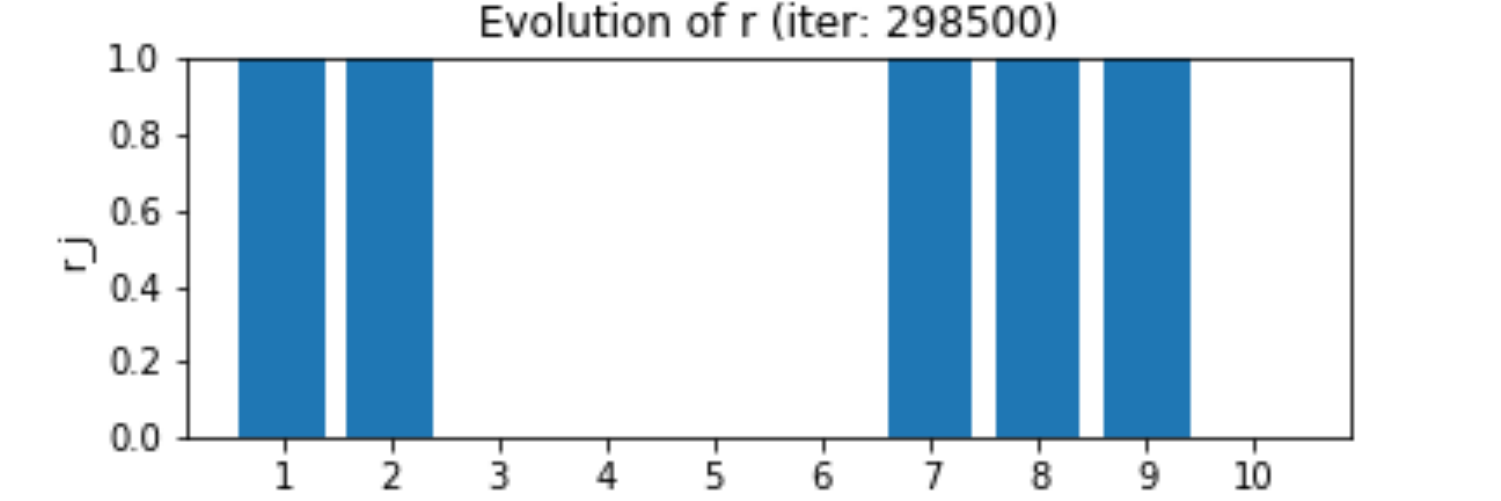} 
\includegraphics[trim = 13.0mm 0mm 12.7mm 5mm, clip, scale=0.313]{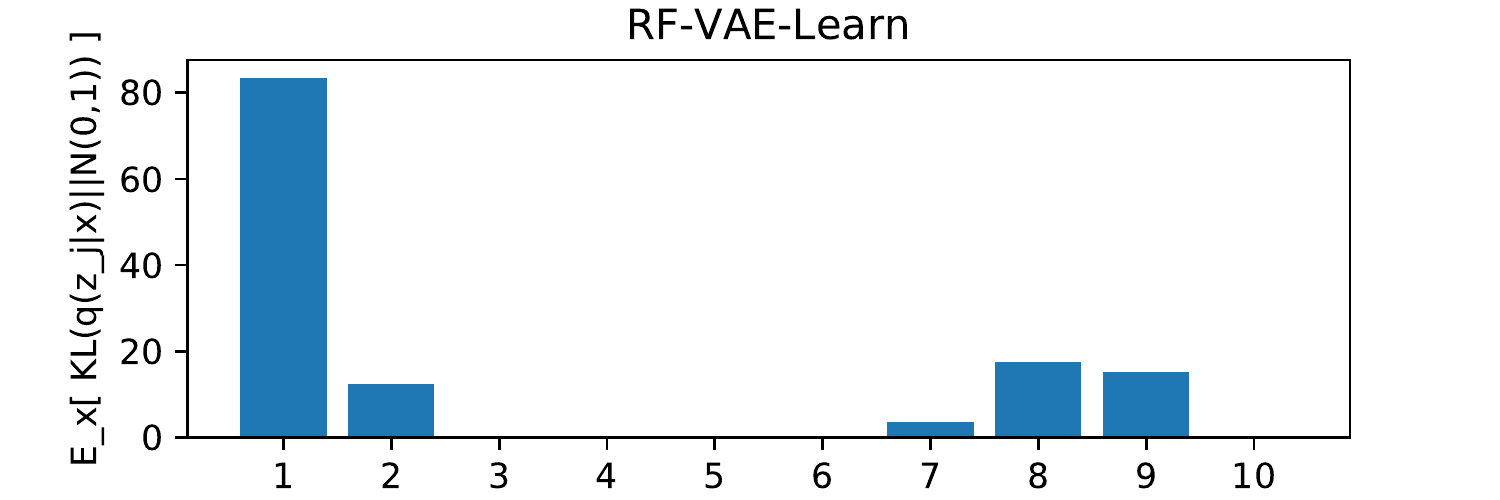}
\end{center}
\vspace{-1.5em}
\caption{RF-VAE on \texttt{Sprites Oval-Only}. (Left) Learned relevance vector ${\bf r}$. (Right) Expected prior KLs. 
}
\label{fig:rfvae_learned_dsprites_oval}
\vspace{-1.0em}
\end{figure}

The disentanglement scores are summarized in~\autoref{tab:dsprites_oval_metrics}, where now both RF-VAE-0 and RF-VAE yield nearly perfect performance in Metric I and II.  Other models' scores fall significantly below.
%
Our RF-VAE identifies five latent dimensions as relevant, \autoref{fig:rfvae_learned_dsprites_oval}, spreading the rotation across two latents. The images generated by traversal shown in~\autoref{fig:rfvae_traverse_dsprites_oval} qualitatively align with the reported scores, indicating the ability of our RF-VAE to recover the disentangled factors.

\subsubsection{Teapots}\label{sec:teapots}

The dataset contains 200,000 images of a teapot across 
five different sources of variation: azimuth, elevation, and the color of the teapot object. Again we choose ten dimensions for the latent vector ${\bf z}$. 

\begin{figure}
\begin{center}
\includegraphics[trim = 8mm 0mm 10mm 4mm, clip, scale=0.315]{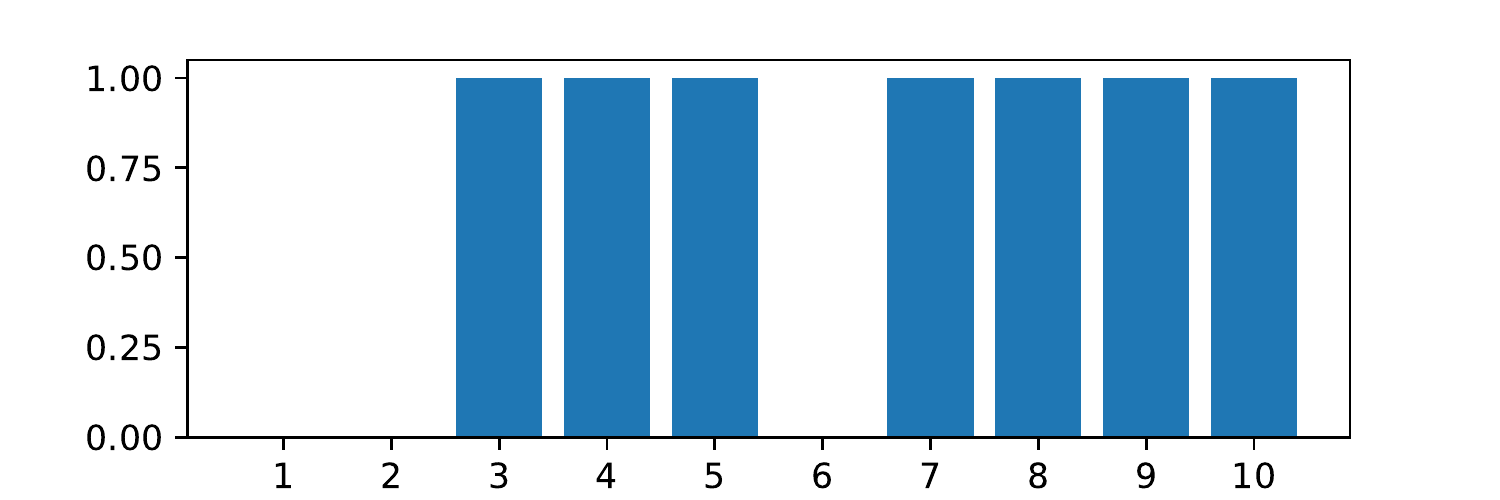} 
\includegraphics[trim = 14mm 0mm 15mm 4mm, clip, scale=0.315]{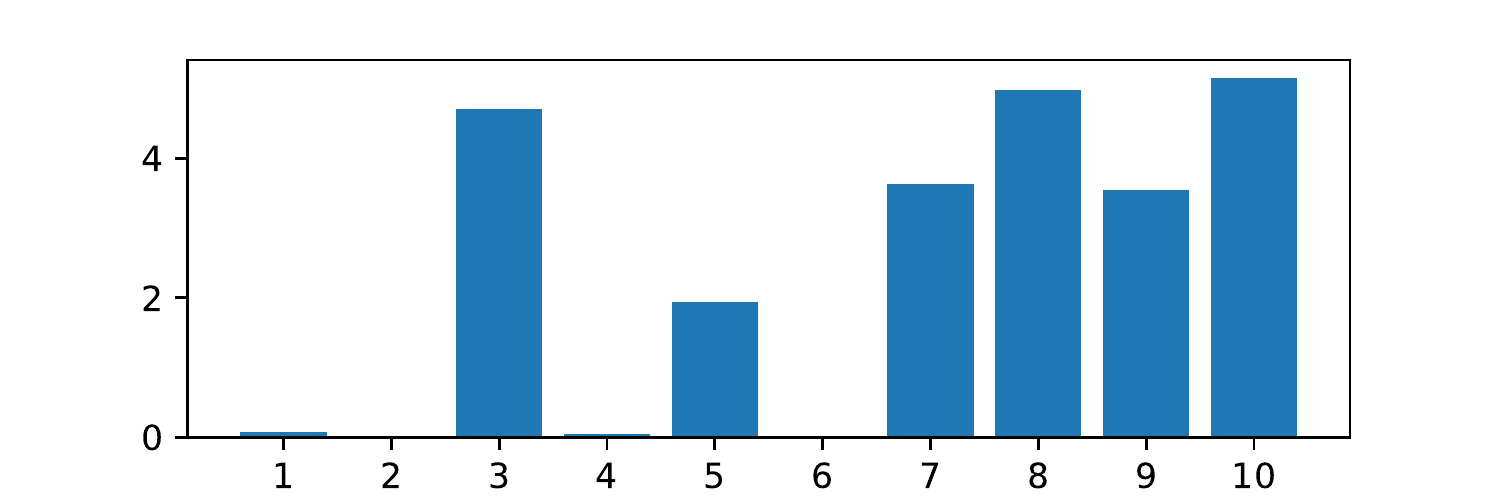}
\end{center}
\vspace{-1.3em}
\caption{RF-VAE on \texttt{Teapots}. (Left) Learned relevance vector ${\bf r}$. (Right) Expected prior KLs for individual dimensions. 
}
\label{fig:rfvae_learned_teapots}
\vspace{-1.5em}
\end{figure}

The disentanglement scores are summarized in ~\autoref{tab:teapots_metrics}. Since the factor labels provided in the dataset are real and uniquely valued, to evaluate metric I and II, we discretized them into ten bins. Interestingly, RF-VAE-0, which uses the knowledge of the number of true factors (5), performed worse than competing models. One possible explanation is that the model architecture may not be flexible enough to represent the full variation in the entire data\footnote{This is a conjecture. Note that a more complex ResNet used in~\cite{williams18} failed in a similar manner. More rigorous failure analysis will be conducted in future work.}. This is supported in the fact that RF-VAE, which identified $7 > 5$ relevant factors, \autoref{fig:rfvae_learned_teapots}, attains higher performance. However, as shown in~\autoref{fig:rfvae_traverse_teapots}, two ($z_4, z_5$) of the relevant dimensions are not indicative of the variations, possibly acting as a conduit to the decoding image generation process. 
It is worth noting that $\beta$-VAE performs poorly on this dataset, possibly due to its known drawback: it over-emphasizes the regularization of matching $p({\bf z})$ and $q({\bf z}|{\bf x})$ across all ${\bf x}$, which effectively makes ${\bf x}$ and ${\bf z}$ less co-dependent, resulting in the learned representation that carries less information from the input, as noted in~\cite{fixing_elbo}. 

\section{Conclusions}\label{sec:conclusion}

In this paper we introduced a new VAE model family, whose goal is to learn succinct, disentangled representations of a dataset.  Starting with the TC-based disentanglement constraint that emphasizes independence of latent factors, we solved the key issue of prior approaches which lack the ability to discern relevant from irrelevant, nuisance factors.  Using a relevance factor formulation, we proposed a learning approach to automatically identify the two groups of factors from data, simultaneously with the process of VAE model parameter estimation.  We also provided a theoretical analysis that underpins this approach.  

Our empirical evaluation on benchmark datasets relied on both quantitative disentanglement scores, including the newly proposed Metric II, and the qualitative traversal analysis.  On most benchmarks, our model exhibited the desired ability to separate relevant from nuisance factors, which helped the model more accurately identify the true, independent sources of data variation. This was particularly evident in the case of synthetic, controlled datasets, where our model outperformed current state-of-the-art.  However, we observed that the existence of discrete factors of low cardinality could adversely affect the model's performance, sometimes resulting in entanglement of the discrete factor with a continuous source of variation.  We posit that these adverse effects could be alleviated by extending the proposed model to a more general, hybrid factor framework.  In the case of a challenging real-world dataset, Celeb-A, our model was able to quantitatively and qualitatively identify many of the key attributes of variation, despite their discrete structure and possible co-dependencies.

%





\end{document}